\documentclass[journal]{IEEEtran}

\usepackage{amsmath,amssymb,amsthm}

\usepackage{hyperref}
\usepackage{stmaryrd}
\usepackage{cleveref}
\usepackage{mathtools}

\let\oldReturn\Return
\newcommand{\Return}{\State\oldReturn}

\newcommand{\bx}{\ensuremath{\mathbf{x}}}

\newcommand{\tS}{\tilde{S}}

\newcommand{\tlambda}{\tilde{\lambda}}
\renewcommand{\th}{\tilde{h}}
\newcommand{\tx}{\tilde{x}}

\newcommand{\normed}[1]{\left\lVert #1\right\rVert}

\newcommand{\cG}{\ensuremath{\mathcal{G}}}
\newcommand{\cE}{\ensuremath{\mathcal{E}}}
\newcommand{\cH}{\ensuremath{\mathcal{H}}}
\newcommand{\cV}{\ensuremath{\mathcal{V}}}
\newcommand{\cO}{\ensuremath{\mathcal{O}}}

\newcommand{\cL}{\ensuremath{\mathcal{L}}}

\newcommand{\R}{\ensuremath{\mathbb{R}}}

\newtheorem{definition}{Definition}
\newtheorem{remark}{Remark}
\newtheorem{lemma}{Lemma}
\newtheorem{theorem}{Theorem}
\newtheorem{corollary}{Corollary}

\newtheorem{proposition}{Proposition}
\theoremstyle{definition}

\numberwithin{problem}{subsection}

\usepackage{cite}

\usepackage{algorithmic}
\usepackage{url}
\usepackage{algorithm}
\usepackage{color}
\usepackage{enumitem}
\usepackage{bbding}

\usepackage{flushend}

\usepackage[labelformat=simple]{subcaption}

\usepackage{graphicx}
\usepackage{epstopdf}
\graphicspath{{./Plots/}}

\newcommand{\tbx}{\ensuremath{\tilde{\bx}}}

\newcommand{\hgnnFull}{Hyper-graph Expansion Neural Network}
\newcommand{\hgnn}{HENN}

\newcommand{\checkmk}{\CheckmarkBold}

\newcommand{\cI}{\mathcal{I}}
\newcommand{\cP}{\mathcal{P}}

\newcommand{\bv}{\mathbf{v}}
\newcommand{\by}{\mathbf{y}}
\newcommand{\bz}{\mathbf{z}}

\begin{document}
\title{Stable and Transferable Hyper-Graph Neural Networks}
\author{Mikhail~Hayhoe,
        Hans~Riess,
        Victor M. Preciado,
        and~Alejandro~Ribeiro%
\thanks{M. Hayhoe, V. M. Preciado, and A. Ribeiro are with the Department of Electrical and Systems Engineering, University of Pennsylvania, Philadelphia, PA, 19104 USA e-mail: \texttt{mhayhoe@seas.upenn.edu}.

H. Riess is with the Department of Electrical and Computer Engineering, Duke University, Durham, NC, 27708 USA.}%
}%

\maketitle

\begin{abstract}
We introduce an architecture for processing signals supported on hypergraphs via graph neural networks (GNNs), which we call a \hgnnFull~ (\hgnn), and provide the first bounds on the stability and transferability error of a hypergraph signal processing model. To do so, we provide a framework for bounding the stability and transferability error of GNNs across arbitrary graphs via spectral similarity. By bounding the difference between two graph shift operators (GSOs) in the positive semi-definite sense via their eigenvalue spectrum, we show that this error depends only on the properties of the GNN and the magnitude of spectral similarity of the GSOs. Moreover, we show that existing transferability results that assume the graphs are small perturbations of one another, or that the graphs are random and drawn from the same distribution or sampled from the same graphon can be recovered using our approach. Thus, both GNNs and our \hgnn s (trained using normalized Laplacians as graph shift operators) will be increasingly stable and transferable as the graphs become larger. Experimental results illustrate the importance of considering multiple graph representations in \hgnn, and show its superior performance when transferability is desired.
\end{abstract}


\section{Introduction}

Graph Neural Networks (GNNs) are information processing and learning architectures for signals supported on graphs \cite{bronstein2017geometric,ruiz2021graph}. They have been widely used in practice for problems ranging from text analysis \cite{defferrard2016convolutional} to recommendation \cite{ying2018graph} to control of multi-agent robotic systems \cite{tolstaya2020learning}, among many others \cite{wu2020comprehensive,zhou2020graph}. However, in many practical applications, relationships between entities are not inherently pairwise; examples include brain networks \cite{giusti2016two}, biochemical reactions \cite{klamt2009hypergraphs}, social interactions \cite{kee2013social}, and more. These higher-order relationships are often modeled using every pairwise relationships between group members, but can be more faithfully represented via sets of entities where the cardinality of the set can be greater than two, and the interactions themselves can be nonlinear in nature. Hypergraphs (or higher-order graphs) and simplicial complices are tools for representing these higher-order relationship and have seen use in many applications \cite{schaub2021signal}. Simplicial complices require the collection of node sets (called simplices) to be closed to taking subsets \cite{salnikov2018simplicial}; hence, they can be viewed as hypergraphs with added constraints that in turn provide added structure. While this additional simplicial constraint may be impractical, it endows rich topological structure known as \emph{Hodge theory} (see, e.g., \cite{schaub2020random,schaub2021signal}). The so-called \emph{Hodge Laplacians} and their corresponding eigenvalue spectra can be understood as higher-order spectral analogues of the ordinary graph Laplacian, and have seen success for higher-order graph learning \cite{roddenberry2021principled,hajij2021simplicial,bodnar2021weisfeiler,yang2022simplicial,giusti2022simplicial,barbarossa2020topological}. Building on these concepts, we introduce a hypergraph learning framework called \hgnnFull s (\hgnn s) by combining graph representations of hypergraphs. While other approaches use graph representations for hypergraph signal processing \cite{feng2019hypergraph, gao2022hgnn, bai2021hypergraph, yadati2019hypergcn, dong2020hnhn}, none combine all of the spectral interpretation and Hodge theory via simplicial complices, nonlinear interactions between nodes and hyperedges, and ability to process both node and hyperedge signals.

Informally, transferability is a type of generalization property for graph signal processing architectures which says that, given two graphs that describe the same or similar phenomenon, the graphs should process signals in a similar manner \cite{levie2021transferability}. There have been three main approaches for codifying the similarity of graphs, i.e., that graphs model similar phenomena. The first compares the original graph to a mildly perturbed version \cite{gama2020stability, levie2019transferability, kenlay2021interpretable}, although the notions of perturbation and the measure of transferability differ. The second approach assumes a latent space model by which similar graphs are created. For example, the nodes of similar graphs may belong to the same latent measure space with signals sampled accordingly \cite{levie2021transferability}, or the graphs themselves may be random and drawn from the same distribution \cite{keriven2020convergence}. The third approach assumes similar graphs are obtained from a graphon, which can be understood as the continuous limit objects of sequences of graphs \cite{lovasz2012large}. Transferability bounds using graphons have been explored in both the asymptotic sense \cite{ruiz2020graphon} and the non-asymptotic sense \cite{maskey2021transferability,ruiz2021graph}.

While these approaches describe transferability of GNNs across similar graphs, they may not be useful in practice. The core concept is that GNNs are spectral operators and, hence, should be transferable across graphs with similar spectra. This \emph{spectral similarity} is achieved as a consequence of the assumptions under which the similar graphs are obtained, generated, or sampled. Unfortunately, given two real graphs of interest, it may not be possible to assert that they are small perturbations of one another, or are sampled from the same latent space or graphon. As such, it is of great practical interest to obtain transferability bounds across \emph{arbitrary} graphs, without any assumptions on their origin. To this end, we provide the first GNN transferability bound directly between two arbitrary graphs of the same size, which can be computed before any training has occurred\footnote{As we will show, certain design choices for the GNN architecture will affect its transferability.}. The key tool in our analysis is to explicitly measure the \emph{spectral similarity} \cite{spielman2011spectral,batson2013spectral} of the graphs for which the transferability bound is desired. 

Using our approach, we provide what is to the best of our knowledge the first bound on the transferability performance of neural networks that perform convolutions with signals supported on arbitrary hypergraphs. The key herein is to consider graph representations of hypergraphs (described in \Cref{sec:hypergraphs}), which may have arbitrary structure; hence, previous results on GNN transferability which make assumptions on graph structure may not apply. In contrast, we may apply our results by simply measuring the spectral similarity of the graph representations of the hypergraphs of interest, before any training has occurred. We also explore transferability without direct computation of spectral similarity in the context of small perturbations, random graphs, and graphs sampled from graphons to show that transferability bounds in each of these regimes may be recovered using our approach.

\section{Preliminaries}

\subsection{Graph Neural Networks}\label{sec:GNN}

A \emph{graph} is a tuple $\cG = (\cV, \cE, W)$ with a set of $n$ nodes $\cV$, $m$ edges $\cE \subseteq \cV\times\cV$, and a real edge weighting function $W : \cE \to \R$. We assume throughout that $\cG$ is undirected and connected. The graph $\cG$ can be represented using a matrix $S \in \R^{n\times n}$ which respects its sparsity pattern, i.e., $[S]_{ij} = 0$ whenever $(i,j) \not\in\cE$, with popular examples including the adjacency matrix, graph and random walk Laplacians, and the normalized versions thereof. Since $\cG$ is undirected with real edge weights the matrix representation $S$ is symmetric and diagonalizable, with an orthonormal eigenvector basis $V = [\bv_1,\ldots,\bv_n]$ and eigenvalue matrix $\Lambda = \text{diag}(\lambda_1,\ldots,\lambda_n)$, where the eigenvalues are real and ordered so that $\lambda_1 \leq \cdots \leq \lambda_n$. Node signals are data vectors $\bx = [x_1,\ldots,x_n]^\intercal \in \R^n$ where $x_i$ is associated to node $i$. We often assume individual node signals are scalars, but this can readily be generalized.

To process a signal $\bx$ using the graph $\cG$, we use a matrix representation $S$ as a \emph{graph shift operator} (GSO) via the linear map $\by = S^k\bx$, where the $k$-fold application of $S$ represent local exchanges of information from the signal $\bx$ between a node and its neighbors which are at most a $k$-length path away in the graph $\cG$ \cite{ruiz2021graph}. With this in mind, we define graph filters as polynomials on the GSO below.

\begin{definition}[Graph filter]
A \emph{graph convolutional filter} $H(S)$ with \emph{filter coefficients} $\{h_k\}_{k=0}^\infty$ is defined as
\begin{align}\label{eq:graph_filter}
    H(S) \coloneqq \sum_{k=0}^\infty h_kS^k.
\end{align}
Moreover, the \emph{graph frequency response} of the filter is
\begin{align}\label{eq:frequency_response}
    h(\lambda) \coloneqq \sum_{k=0}^\infty h_k\lambda^k.
\end{align}
We frequently consider filters with an analytic frequency response, i.e., a finite number of coefficients, so that for some $K$, $h_k = 0~\forall k>K$. Moreover, we assume $|h(\lambda)| \leq 1$ for all $\lambda \in [\lambda_1(S), \lambda_n(S)]$ so filters do not amplify signals, which is trivially satisfied via normalization with finite coefficients.
\end{definition}

\noindent In order to improve the representation power of graph filters, in practice they are stacked together with pointwise nonlinearities to create a \emph{graph neural network} (GNN), defined below.

\begin{definition}[Graph neural network]
Graph neural networks are a cascade of $L$ layers of graph filters, each followed by a pointwise nonlinearity. Let each layer have $f_l$ graph signals (or \emph{features}) $\bx_l^1,\ldots,\bx_l^{f_l} \in \R^n$. At layer $l$, we apply $f_l f_{l-1}$ graph filters of the form $H^{ij}_l(S)$ followed by a pointwise (or elementwise) nonlinear function $\sigma : \R \to \R$ to process the $f_{l-1}$ input features into the $f_l$ output features via
\begin{align*}
    \bx_l^i = \sigma\left(\sum_{j=1}^{f_{l-1}}\sum_{k=0}^\infty h_{lk}^{ij}S^k\bx_{l-1}^j \right), \quad i \in \{1,\ldots,f_l\}.
\end{align*}
A graph neural network is the mapping $\Phi(\bx_0; S, H) = \bx_L$. In this work we consider normalized Lipschitz continuous nonlinearities so that $|\sigma(x) - \sigma(y)| \leq |x - y|$, e.g., \emph{ReLU}, \emph{sigmoid}, and \emph{tanh}.
\end{definition}

A property of graph filters which has been shown to be valuable for stability of GNNs \cite{gama2020stability, ruiz2021graph} is the so-called \emph{integral Lipschitz} condition, defined below.

\begin{definition}[Integral Lipschitz]
A filter with frequency response $h$ is \emph{integral Lipschitz} on an interval $\cI$ if there is some $C>0$ such that, for all $\lambda_1,\lambda_2 \in \cI$,
\begin{align}
    |h(\lambda_1) - h(\lambda_2)| \leq C\frac{|\lambda_1 - \lambda_2|}{|\lambda_1 + \lambda_2|/2},
\end{align}
which implies that the derivative of $h$ satisfies $|\lambda h'(\lambda)| \leq C$. We omit the interval $\cI$ when it is clear from context, e.g., for filters that will be applied to some GSOs $S_1,\ldots,S_m$ we have $\cI = \cup_{i=1}^m [\lambda_1(S_i), \lambda_n(S_n)]$.
\end{definition}

Integral Lipschitz graph filters can be arbitrarily discriminative for small eigenvalues, but must become effectively flat for larger eigenvalues. However, by applying nonlinearities to the output of these filters, the portion of the spectrum containing larger eigenvalues may be scattered to the lower portion. In other words, GNNs with integral Lipschitz filters can be both discriminative and stable \cite{gama2020stability}. Given some finite GSO $S$ as well as a filter $H(S)$ with an analytic frequency response and support contained in $[\lambda_1(S), \lambda_n(S)]$, the integral Lipschitz condition is trivially satisfied with 
\begin{align}\label{eq:integral_Lipschitz_constant}
    C = \max\left\{\left|\sum_{k=1}^K h_kk\lambda_{1}(S)^k\right|, \left|\sum_{k=1}^K h_kk\lambda_{n}(S)^k\right|\right\}.
\end{align}
Furthermore, we note that the value of the integral Lipschitz constant in a learning architecture may be affected by adding a penalty term to the loss function used for training.

\section{Hyper-Graph Neural Networks}\label{sec:hypergraphs}

\begin{figure*}
\centering
    \begin{subfigure}[t]{0.15\linewidth}
        \centering
    	\includegraphics[width=0.9\linewidth]{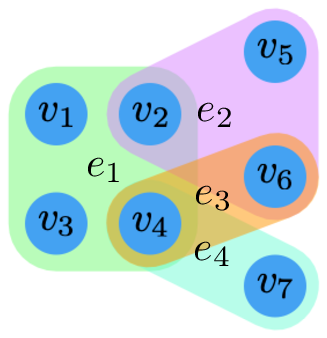}
    	\caption{Hypergraph}
    	\label{fig:hypergraph}
    \end{subfigure}
    ~
    \begin{subfigure}[t]{0.15\linewidth}
        \centering
    	\includegraphics[width=\linewidth]{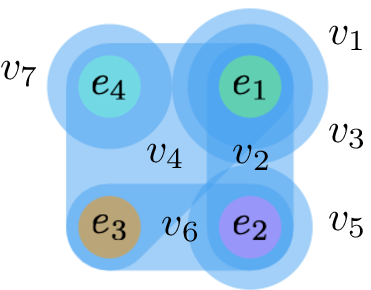}
    	\caption{Dual hypergraph}
    	\label{fig:dual_hypergraph}
    \end{subfigure}
    ~
    \begin{subfigure}[t]{0.15\linewidth}
        \centering
    	\includegraphics[width=0.9\linewidth]{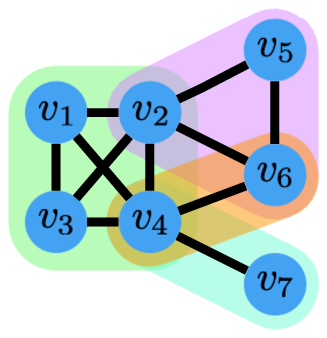}
    	\caption{Clique expansion}
    	\label{fig:clique_expansion}
    \end{subfigure}
    ~
    \begin{subfigure}[t]{0.15\linewidth}
        \centering
    	\includegraphics[width=0.8\linewidth]{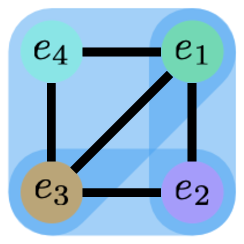}
    	\caption{Line graph}
    	\label{fig:line_graph}
    \end{subfigure}
    ~
    \begin{subfigure}[t]{0.16\linewidth}
        \centering
    	\includegraphics[width=0.9\linewidth]{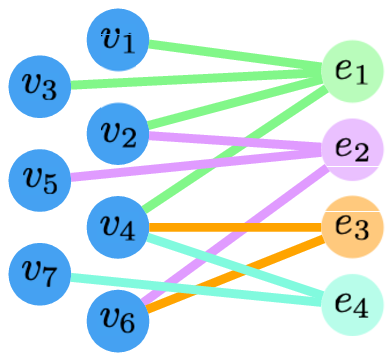}
    	\caption{Bipartite expansion}
    	\label{fig:bipartite_expansion}
    \end{subfigure}
    ~
    \begin{subfigure}[t]{0.15\linewidth}
        \centering
    	\includegraphics[width=0.9\linewidth]{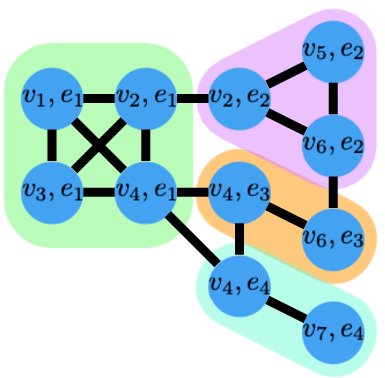}
    	\caption{Star expansion}
    	\label{fig:star_expansion}
    \end{subfigure}
    \caption{Example hypergraph and graph representations, none of which faithfully represent the original hypergraph on their own. The clique expansion in \ref{fig:clique_expansion} cannot distinguish the real hyperedge $e_2$ from the clique $\{v_4, v_6, v_7\}$. The line graph in \ref{fig:line_graph} cannot distinguish the dual hyperedge of $v_4$ from the clique $\{e_1, e_2, e_3\}$. The bipartite expansion in \ref{fig:bipartite_expansion} has heterogeneous nodes and a different notion of signal processing. Finally, the star expansion in \ref{fig:star_expansion} adds many new nodes for each hyperedge intersection, which is impractical for dense hypergraphs and removes the interpretability of node convolutions.}
    \label{fig:hg_expansions}
\end{figure*}

\begin{table}[ht]
    \centering
    \caption{Notation for a hypergraph with $n$ nodes and $m$ hyperedges}
    \label{tab:my_label}
    \begin{tabular}{|c|l|}
        \hline
        Symbol & Meaning \\
        \hline
        $A \in \R^{n\times n}$ & Adjacency matrix \\
        $B \in \R^{n\times m}$ & Node-hyperedge incidence matrix \\
        $D_v \in \R^{n\times n}$ & Diagonal node degree matrix \\
        $D_e \in \R^{m\times m}$ & Diagonal hyperedge size matrix \\
        $D_{ee} \in \R^{m\times m}$ & Diagonal hyperedge intersection count matrix \\
        $W \in \R^{m\times m}$ & Diagonal hyperedge weight matrix \\
        \hline
    \end{tabular}
\end{table}

A \emph{hypergraph}, or higher-order graph, is a tuple $\cH = (\cV, \cE, W)$ with a set of $n$ nodes $\cV$ and a collection of $m$ hyperedges $\cE \subseteq 2^{\cV}$, where $2^{\cV}$ is the powerset of $\cV$, i.e., the collection of all subsets of $\cV$. Thus the hyperedges are arbitrarily-sized sets of nodes, to which we ascribe some hyperedge weighting function $W: \cE \to \R$. With an abuse of notation, we stack the hyperedge weights into a diagonal matrix $W$. If all hyperedges have cardinality two, we recover the ordinary definition of a graph. Hypergraphs are commonly represented as matrices via the node-hyperedge \emph{incidence matrix} $B \in \R^{n\times m}$, where
\begin{align}\label{eq:incidence_matrix}
    [B]_{ij} = \begin{cases}
        1, &\text{node $i$ is in hyperedge $j$}, \\
        0, &\text{otherwise}.
    \end{cases}
\end{align}
The \emph{energy} of a hypergraph node signal $x$ on $\cH$ is defined as
\begin{align}\label{eq:hg_energy}
    Q_{\cH}(x) \coloneqq \sum_{e \in \cE} \max_{i,j \in e}(x_i - x_j)^2.
\end{align}
The \emph{hypergraph Laplacian} is then defined as the gradient of the energy functional, i.e.,
\begin{align}\label{eq:hg_laplacian}
    \cL_{\cH}(x) \coloneqq \frac{1}{2}\nabla Q_{\cH}(x).
\end{align}
The hypergraph energy \eqref{eq:hg_energy} is a generalization of the graph energy of a signal \cite{batson2013spectral} defined for a graph $\cG = (\cV, \cE_\cG, W)$ as
\begin{align}
    Q_\cG(x) = \sum_{(i,j) \in \cE_\cG} (x_i - x_j)^2 = x^\intercal BB^\intercal x = x^\intercal \cL_\cG x,
\end{align}
where $B$ is the node-edge incidence matrix, and $\cL_\cG$ is the graph Laplacian. In contrast to many graph Laplacians, however, $\cL_\cH$ is a nonlinear operator. As such, its resultant diffusions are not easily-approximated with any linear graph Laplacian. Indeed, in contrast to a graph diffusion between nodes, such a hypergraph operator diffuses the signal $x$ across both the nodes and hyperedges.

A common technique for processing hypergraph signals is to use graph representations \cite{feng2019hypergraph, gao2022hgnn, bai2021hypergraph, yadati2019hypergcn, dong2020hnhn}. While many representations are used in practice \cite{schaub2021signal}, the most common are the clique expansion, line graph, star expansion, and bipartite expansion (see \Cref{fig:hg_expansions}). The clique expansion is the graph generated by replacing each hyperedge by a clique, and the line graph is the clique expansion of the dual hypergraph (wherein the roles of nodes and hyperedges are reversed).
The star expansion makes node-hyperedge pairs, and places edges between these pair-nodes if they shared a node or hyperedge in the hypergraph. The bipartite expansion partitions the nodes and hyperedges, and places edges between them based on hyperedge inclusions in the hypergraph.

In principle, any graph representation(s) may be used in a hypergraph signal processing framework. However, in practice, the star expansion cannot represent large and/or dense hypergraphs due to its size. Moreover, the bipartite expansion loses the meaning of convolutions via applications of the graph shift operator, since the node set and, hence, the node signals are heterogeneous. In contrast, the clique expansion and line graph are homogeneous and of reasonable size; moreover, they are intimately related to the theory of simplicial complices. Indeed, since simplicial complices require closure under taking subsets, we can build a simplicial complex from a hypergraph by taking the original hyperedges and adding any missing subsets of these hyperedges as simplices. The clique expansion is then simply the 1-skeleton of this simplicial representation of the hypergraph, i.e., it includes only the $0$-simplices (nodes) and $1$-simplices (edges) and throws away all higher-order simplices. The line graph can also be seen as the 1-skeleton of the nerve complex constructed from the hypergraph \cite{grunbaum1970nerves}, which is equivalent to the simplicial complex constructed from the dual hypergraph. Finally, we remark that the hypergraph Laplacian in \eqref{eq:hg_laplacian} is a nonlinear operator that diffuses information across \emph{both} the nodes and hyperedges and, hence, cannot be well-approximated via the linear graph Laplacian of one representation alone. For these reasons we introduce \hgnn~below, which combines convolutions using the clique expansion and line graph.

\begin{table*}[t]
    \centering
    \caption{Comparison of hypergraph signal processing approaches which use graph representations, including their connections with higher-order spectral theory, the nonlinear hypergraph Laplacian \eqref{eq:hg_laplacian}, simplicial complices (SCs), and whether they can process both node and hyperedge signals.}
    \label{tab:hg_convolutions}
    
    \begin{tabular}{|c|c|c|c|c|c|}
        \hline
        Architecture & GSOs & Spectral & HG Laplacian & SCs & Node \& hyperedge signals\\
        \hline
        \hgnn~(ours) & $D_v^{-1/2}BWB^\intercal D_v^{-1/2}$ \& $D_{ee}^{-1/2}B^\intercal BW D_{ee}^{-1/2}$ & \checkmk & \checkmk & \checkmk & \checkmk \\
        \hline
        HGNN \cite{feng2019hypergraph} & $D_v^{-1/2}BWD_e^{-1}B^\intercal D_v^{-1/2}$ & \checkmk & & \checkmk & \\
        \hline
        HGNN+ \cite{gao2022hgnn} & $D_v^{-1}BWD_e^{-1}B^\intercal$ & \checkmk & & & \checkmk \\
        \hline
        HyperAtten \cite{bai2021hypergraph} & $D_v^{-1/2}\tilde{B}WD_e^{-1}\tilde{B}^\intercal D_v^{-1/2}$ & \checkmk & & &  \\
        \hline
        HyperGCN \cite{yadati2019hypergcn} & $\tilde{D}_v^{-1/2}\tilde{B}WD_e^{-1}\tilde{B}^\intercal \tilde{D}_v^{-1/2}$ & & \checkmk & & \\
        \hline
        HNHN \cite{dong2020hnhn} & $D_e^{-1}B^\intercal D_v $ \& $D_v^{-1} B D_e$ & \checkmk & & & \checkmk \\
         \hline
    \end{tabular}
\end{table*}

\begin{definition}[\hgnn]
A \hgnnFull, or \hgnn, $\Phi(\cdot; S_c,\Theta_c,S_l,\Theta_l)$ is composed of cascades of clique expansion layers of the form
\begin{align}
    X_v^{l+1} &= \sigma\left(\sum_{k=0}^K\left(D_v^{-1/2}BWB^\intercal D_v^{-1/2}\right)^k X_v^{l}H_v^{k,l}\right),
\end{align}
and line graph layers of the form
\begin{align}
    X_e^{l+1} &= \sigma\left(\sum_{k=0}^K\left(D_{ee}^{-1/2}B^\intercal BW D_{ee}^{-1/2}\right)^k X_e^{l}H_e^{k,l}\right),
\end{align}
where $D_{ee}$ is the hyperedge degree matrix from the line graph. We pool between these types of layers using max-pooling according to node-hyperedge membership, i.e., based on the entries of $B$, following the definition of the nonlinear hypergraph Laplacian in \eqref{eq:hg_laplacian}.
\end{definition}

There are several existing approaches for learning with hypergraphs via graph convolutions. We summarize their methodologies, benefits, and particular GSOs in our notation in \Cref{tab:hg_convolutions}. HGNN \cite{feng2019hypergraph} performs convolutions using a hyperedge-normalized and weighted clique expansion of the hypergraph via the incidence matrix. 
HGNN+ \cite{gao2022hgnn} performs a convolution using the weighted and normalized incidence matrices, but the resulting convolution is interpreted as convolving the hyperedge signals via $WD_e^{-1}B^\intercal$ and then a node convolution via $D_v^{-1}B$. 
This process may include multi-modal nodes and/or hyperedges, in which case the separately learned embeddings are joined together (stacked, pooled, etc.). 
HNHN \cite{dong2020hnhn} uses the incidence matrices as GSOs in a two-phase approach, similarly to HGNN+ with an extra nonlinearity added between the hyperedge and node convolutions. 
HyperAtten \cite{bai2021hypergraph} adapts HGNN by adding an attention module to adaptively learn the weights of the incidence matrix $\tilde{B}$, in cases where the hyperedges and nodes are from the same domain (e.g., hyperedges are formed from nearest neighbours of nodes). 
HyperGCN \cite{yadati2019hypergcn} builds a restricted clique expansion and performs convolutions. Instead of full cliques for each hyperedge, in each epoch, for each graph filtering layer and each hyperedge $e$ it includes only the edge $(i,j) = \arg\max_{i,j\in e}\normed{x^l_i - x^l_j}_2$ between the nodes with the largest pairwise signal difference, following \eqref{eq:hg_laplacian}. Edges between these two nodes and the other nodes in the hyperedge may also be included using much smaller weights. Note that the resulting GSO will depend on the node signals, of which there may be many in a dataset.

\subsection{Spectral similarity}\label{sec:spectral_similarity}

We quantify the similarity of connected, undirected graphs with the same number of nodes, including graph representations of hypergraphs, by measuring the similarity of the spectra of their graph shift operators (GSOs), which we assume to be symmetric and positive semi-definite. We stress that a graph admits many different shift operators, such as the normalized graph Laplacian, and a hypergraph admits many graph representations, such as the clique expansion and line graph. While our notion of similarity will be dependent on which graph representations and/or GSOs we consider, this is an appropriate measure for similarity when signals are being processed by these particular GSOs.

\begin{definition}[Spectral similarity \cite{spielman2011spectral}]\label{def:spectral_similarity}
The symmetric and positive semi-definite matrices $S \in \R^{n\times n}$ and $\tS \in \R^{n\times n}$ are called \emph{$\epsilon$-spectrally similar} if $(1-\epsilon)S \preceq \tS \preceq (1+\epsilon)S$, i.e.,
\begin{align}\label{eq:spectral_similarity}
    (1-\epsilon)x^\intercal Sx \leq x^\intercal \tS x \leq (1+\epsilon)x^\intercal S x \quad\forall x\in\R^n,
\end{align}
which implies,
\begin{align}
    (1-\epsilon)\lambda_i(S) \leq \lambda_i(\tS) \leq (1 \!+\! \epsilon)\lambda_i(S) ~\forall i\in\!\{1,\ldots,n\}.
\end{align}
\end{definition}
For some arbitrary graphs $\cG$ and $\tilde{\cG}$ with $n$ nodes and GSOs $S$ and $\tS$, respectively, the coefficient of spectral similarity can be computed to precision $\kappa$ in time polynomial in $n$ and $\log(1/\kappa)$ \cite{batson2013spectral}, for example via semi-definite programming. If the GSOs of interest are diagonally dominant (e.g., graph Laplacians), we may instead employ linear programming to greatly increase scalability \cite{ahmadi2019dsos}. Note also that the multiplicity of the eigenvalue zero must be the same (possibly both zero) for $S$ and $\tS$, which will be the case for most GSOs of connected graphs, such as the normalized Laplacian.

The coefficient of spectral similarity, $\epsilon$, will be the quantity by which we will provide bounds on the transferability of GNNs between two arbitrary graphs of the same size. This coefficient may \emph{always} be measured between the GSOs of arbitrary graphs with the same number of nodes, but it may in general be quite large. However, for a GNN to be transferable we require only that the output is close for similar graphs. Indeed, if the output of a GNN was similar when considering graphs that are not spectrally similar, the GNN would have poor ability to discriminate.

Finally, we mention that explicit computation of the spectra may be prohibitively expensive for very large graphs. Thankfully, if certain conditions on the graphs of interest are satisfied, we may still be able to bound and/or compute the spectral similarity. We explore the case of random graphs in \Cref{sec:RG_tf}, graphs sampled from graphons in \Cref{sec:graphon_tf}, and small perturbations in \Cref{sec:perturbation_similarity}, further showing the generality of our transferability bounds in the context of previous works.

\section{Transferability via spectral similarity}\label{sec:transferability}

To claim that an architecture is transferable, we need to show that using similar graphs to process signals produces similar results. In this paper, we will investigate transferability of graph filters, graph neural networks, and hypergraph neural networks that use graph representations. Since these tools are permutation equivariant \cite{gama2020stability}, we need not be concerned with differences in node labelings between the graphs $\cG$ and $\tilde{\cG}$. To that end, given some signal processing architecture $\Phi$ along with GSOs $S$ and $\tS$, we wish to examine the quantity
\begin{align}
\begin{split}\label{eq:phi_diff}
    &\normed{\Phi(\tS) - \Phi(S)}_{\cP} \\
    &\quad\coloneqq \min_{P \in \cP}\max_{\bx\in\R^n : \normed{\bx}_2=1}\normed{\Phi(\bx; P^\intercal \tS P) - \Phi(\bx; S)}_{2},
\end{split}
\end{align}
which is referred to as the \emph{distance modulo permutation} \cite{gama2020stability}. Here the set of all permutation matrices is denoted $\cP \coloneqq \left\{P \in \{0,1\}^{n\times n} : P\mathbf{1} = P^\intercal\mathbf{1} = \mathbf{1}\right\}$. Note that when $\Phi$ is linear, we may substitute the operator norm and ignore the unit-norm signal $\bx$. If \eqref{eq:phi_diff} is small, then the way $\Phi$ processes signals with $S$ is similar to the way it processes signals using $\tS$ in a worst-case sense, regardless of any differences in the node labeling. In particular, if \eqref{eq:phi_diff} is small whenever $S$ and $\tS$ are $\epsilon$-spectrally similar with small $\epsilon$, then $\Phi$ has the transferability property. We formalize this intuition below.

\begin{proposition}\label{prop:spectral_similarity_bound}
For $\epsilon$-spectrally similar symmetric GSOs $S$ and $\tS$ and an integral Lipschitz filter with constant C, the operator difference modulo permutation between the filters $H(S)$ and $H(\tS)$ satisfies
\begin{align*}
    \normed{H(\tS) - H(S)}_{\cP} \leq C\epsilon + \cO(\epsilon^2).
\end{align*}
Moreover, if the filter applies only one shift operation with bias so $H(S) = h_0 I + h_1 S$, then $\normed{H(\tS) - H(S)}_{\cP} \leq C\epsilon$.
\end{proposition}
\begin{proof}
See \Cref{app:pf_spectral_similarity_bound}.
\end{proof}

\noindent As discussed in \Cref{sec:GNN}, GNNs are cascading layers of graph filters passed through pointwise nonlinearities. Thus we arrive at our main result, which is a bound on the transferability of graph neural networks based on spectral similarity.

\begin{theorem}\label{thm:GNN_bound}
Given $\epsilon$-spectrally similar GSOs $S$ and $\tS$ and a GNN $\Phi(\cdot; S,\Theta)$ with normalized Lipschitz nonlinearities and $L$ layers with $f$ features, each with filters that have unit operator norm and are $C$-integral Lipschitz, then
\[
\normed{\Phi(\cdot; S,\Theta) - \Phi(\cdot; \tS,\Theta)}_\cP \leq CLf^L\epsilon + \cO(\epsilon^2).
\]
\end{theorem}
\begin{proof}
See \Cref{app:pf_GNN_bound}.
\end{proof}

\Cref{thm:GNN_bound} is not surprising; informally, since GNNs are spectral operators, our result says that their actions on any signal $x\in\R^n$ for graphs with similar spectra will be similar. Indeed, with this result in hand, the task of GNN transferability is reduced to measuring the difference of graph spectra. While this bound depends on the design choices of the architecture (such as the number of features and layers), it is independent of the learned parameters of the GNN, $\Theta$, excepting possibly the filters' integral Lipschitz constant $C$. Indeed, if the filters satisfy $|h(\lambda)| \leq 1$ and the integral Lipshitz constant is constrained\footnote{The former may be achieved via normalization during training, and the latter by adding a penalty term to the training loss, such as a log-barrier function on the integral Lipschitz constant, which may be computed via \eqref{eq:integral_Lipschitz_constant}.}, we may obtain this bound on the transferability error between \emph{any arbitrary graphs} $\cG$ and $\tilde{\cG}$ before any training has taken place. In practice, computing the spectrum of a matrix takes at most $O(n^3)$ time, and the coefficient of spectral similarity can then be computed via  \eqref{eq:spectral_similarity_coefficient}. For large graphs this may be computationally expensive; thus, in \Cref{sec:random_transferability,sec:stability} we provide bounds for spectral similarity in many regimes of practical interest. Thus, transferability is entirely characterized by parameters of the architecture (the integral Lipschitz constant, number of features, and number of layers), and the spectral similarity between the graphs of interest. Naturally, worse spectral similarity results in looser transferability bounds, as does a larger integral Lipschitz constant $C$ and more features $f$ or convolutional layers $L$. However, larger $C$, $f$, and $L$ suggest enhanced discriminability, since the GNN can produce sharper filters as $C$ is larger, and more of those filters can be composed as $f$ and $L$ grow. Together, these insights pose a tradeoff between transferability and discriminability; if a GNN is indifferent to large differences in graph spectra, it cannot also treat very similar graphs differently.

We remark that in practice the filters may not be normalized, the integral Lipschitz constants may differ, and each layer may have a different number of features. We explicitly compute our transferability bound in this context in \Cref{app:pf_GNN_bound}.

\section{Transferability of Hyper-Graph Neural Networks}

In this section we will provide what is, to the best of our knowledge, the first bound on the transferability performance of neural networks that perform convolutions with signals supported on arbitrary hypergraphs. The key herein is to consider graph expansions of hypergraphs (described in \Cref{sec:hypergraphs}), which may have arbitrary structure; hence, previous results on GNN transferability which make assumptions on graph structure may not apply. In contrast, we may use our results by simply measuring the spectral similarity of the graph expansions of the hypergraphs of interest.

Consider a hypergraph  $\cH$ and another (similar) hypergraph $\tilde{\cH}$. We may consider $\tilde{\cH}$ as a perturbation of $\cH$, analogously to \Cref{sec:perturbation_similarity}, or it may simply be a related hypergraph that models similar phenomena. Furthermore, consider any hypergraph signal processing framework $\Phi(\cdot;\{S_i,\Theta_i\}_{i=1}^r)$ comprised of graph filtering layers for $r$ graph representations of the original hypergraph. The particular GSOs of the graph representations that are used by many such hypergraph learning frameworks are listed in \Cref{tab:hg_convolutions}; note that most consider only one GSO. By computing the spectral similarities of these graph representations of $\cH$ and $\tilde{\cH}$, the result below allows us to understand how similar the output of the hypergraph learning framework $\Phi(\cdot;\{S_i,\Theta_i\}_{i=1}^r)$ will be when applied to the GSOs $\tS_r,\ldots,\tS_r$.

\begin{theorem}\label{thm:hg_transferability}
Consider two hypergraphs $\cH$ and $\tilde{\cH}$ and $r$ graph representations with GSOs $\{S_i\}_{i=1}^r$ and $\{\tS_i\}_{i=1}^r$, respectively, such that $S_i$ and $\tS_i$ are $\epsilon_i$-spectrally similar. If the hypergraph learning framework $\Phi(\cdot;\{S_i,\Theta_i\}_{i=1}^r)$ has a normalizing pooling function between graph representations with normalized Lipschitz nonlinearities, with $L_i$ layers having $f_i$ features for graph representation $i$, each with filters that have unit operator norm and are $C$-integral Lipschitz, then
\begin{align*}
    &\normed{\Phi(\cdot;\{S_i,\Theta_i\}_{i=1}^r) - \Phi(\cdot;\{\tS_i,\Theta_i\}_{i=1}^r)} \\
    &\quad\leq \sum_{i=1}^r C L_i \epsilon_i \prod_{j=1}^r f_j^{L_j} + \cO(\epsilon_1^2 + \cdots + \epsilon_r^2).
\end{align*}
\end{theorem}
\begin{proof}
Similar to \Cref{thm:GNN_bound} with potentially different GSOs in each layer; see \Cref{app:pf_GNN_bound}.
\end{proof}

Since we make no assumptions on the structure of the hypergraphs and, hence, their graph representations beyond connectedness, \Cref{thm:hg_transferability} provides a transferability bound for \emph{any} hypergraph learning framework which uses graph representations, including all those in \Cref{tab:hg_convolutions}. This result also lends intuition to the formation of \hgnn. From the connection of the clique expansion and line graph with simplicial complices, we know these graph representations are associated with the higher-order spectral theory of Hodge Laplacians \cite{schaub2020random,schaub2021signal}. Together with the results of \Cref{sec:stability}, this suggests that \hgnn~will be stable to structural perturbations in the hypergraph, which is not a statement that can be asserted for the other hypergraph learning methods described earlier. Hence, for two related hypergraphs, \hgnn~will be \emph{both} stable and transferable if the GSOs of their clique expansions and line graphs are spectrally similar.

\section{Transferability between random graphs}\label{sec:random_transferability}

In this section we show that two graphs of the same size drawn from an appropriate random graph distribution will be spectrally similar with a coefficient $\epsilon$ that shrinks as $n$ grows. In other words, random graphs become \emph{more} spectrally similar as they grow larger. By combining this notion with \Cref{thm:GNN_bound}, we obtain results in agreement with works that investigate the transferability of GNNs applied to large random graphs \cite{keriven2020convergence, keriven2021universality} and convergent graph sequences\cite{maskey2021transferability, ruiz2020graphon}.

For practicality and to build intuition throughout this section we will assume that all graph shift operators are normalized Laplacians. However, we stress that this is \emph{not} a requirement of our approach, and any GSO that satisfies the conditions we provide can be used in practice. We begin with a general result which provides the conditions under which a family of random graphs will produce GSOs that grow more spectrally similar as they grow larger, with high probability.

\begin{proposition}\label{prop:RG_similarity}
Let $\{\cG_n\}_{n=1}^\infty$ and $\{\tilde{\cG}_n\}_{n=1}^\infty$ be two independent sequences of graphs drawn from the same family of random graph distributions so that $\cG_n$, $\tilde{\cG}_n \sim P_\cG(n)$ for each $n$, with graph shift operators $S_n$ and $\tS_n$, respectively, having eigenvalues $\{\lambda_i\}_{i=1}^n$ and $\{\tlambda_i\}_{i=1}^n$. Assume the following:
\begin{enumerate}[label=(\emph{A\arabic*})]
    \item\label{ass:zero_multiplicity} Multiplicity of zero: the zero eigenvalue has almost surely constant multiplicity indepedent of $n$ (potentially zero);
    \item\label{ass:spectral_gap} Bounded spectral gap: there exists $c>0$ independent of $n$ such that $|\lambda_i| \geq c$ almost surely for all non-zero eigenvalues;
        \item\label{ass:concentration} Concentration: given $\epsilon_c>0$ and $\delta_c>0$ there exist some values $\gamma_i$, $i\in\{1\ldots,n\}$, such that for large enough $n$,
    \begin{align}
        P\left(|\lambda_i - \gamma_i| < \epsilon_c, ~\forall i\in\{1,\ldots,n\}\right) > 1 - \delta_c.
    \end{align}
\end{enumerate}
Then for any $\epsilon > 0$ and $\delta > 0$, there exists $N = N(\epsilon,\delta)$ such that for any $n \geq N$,
\begin{align}\label{eq:rigidity}
    P\!\left((1\!-\!\epsilon)\lambda_i < \tlambda_i < (1\!+\! \epsilon)\lambda_i, ~\forall i\in\{1,\ldots,n\}\!\right) \!>\! 1 \!-\! \delta.
\end{align}
In other words, for appropriate graph shift operators of large random graphs whose eigenvalues concentrate, the coefficient of spectral similarity converges in probability to $0$ as $n\to\infty$.
\end{proposition}
\begin{proof}
See \Cref{app:pf_RG_similarity}.
\end{proof}

This proposition applies to a large class of random graphs and corresponding shift operators. Assumption \ref{ass:zero_multiplicity} is necessary for spectral similarity to hold but is trivially satisfied by the normalized Laplacian of a connected graph. As we will show in \Cref{sec:RG_tf}, \ref{ass:spectral_gap} and \ref{ass:concentration} are satisfied by the normalized Laplacian of many families of random graphs including the Erd\"os-R\'enyi and Chung-Lu models, as well as random power-law graphs with given expected degree sequences and large enough minimum expected degree \cite{chung2004spectra}. Furthermore, we will show in \Cref{sec:graphon_tf} that graphs sampled from the same well-structured graphon will satisfy all of these conditions and, hence, become arbitrarily spectrally similar as their size grows larger. The implications of these results in terms of transferability of GNNs for random graph families is summarized below.

\begin{theorem}\label{thm:RG_TF_bound}
Let $\{\cG_n\}_{n=1}^\infty$ and $\{\tilde{\cG}_n\}_{n=1}^\infty$ be two independent sequences of random graphs with graph shift operators $\{S_n\}_{n=1}^\infty$ and $\{\tS_n\}_{n=1}^\infty$, respectively, such that $\cG_n, \tilde{\cG}_n \sim P_{\cG}(n)$, and let $P_\cG(n)$ satisfy \ref{ass:zero_multiplicity}-\ref{ass:concentration} for all $n$. Then, for a sequence of GNNs $\{\Phi(\cdot; S_n,\Theta)\}_{n=1}^\infty$ trained on the GSOs $S_n$ with normalized Lipschitz nonlinearities, each with the same number of layers and features, as well as integral Lipschitz filters that have unit operator norm,
\begin{align*}
    \normed{\Phi(\cdot; S_n,\Theta) - \Phi(\cdot; \tS_n,\Theta)}_\cP \to 0,
\end{align*}
in probability as $n \to \infty$.
\end{theorem}
\begin{proof}
Direct result of \Cref{thm:GNN_bound} and \Cref{prop:RG_similarity}.
\end{proof}

The statement above has far-reaching implications. For example, GNNs which use the normalized Laplacian as a GSO will exhibit \emph{better} transferability properties as the size of the original training graph grows, with high probability. This agrees with previous results on GNN transferability in the regime of large random graphs \cite{keriven2020convergence, keriven2021universality}, as well as the regime of convergent random graph sequences \cite{ruiz2020graphon}. In particular, these results suggests that for a family of large enough random graphs, when using the normalized Laplacian as a GSO, it suffices to train on one graph realization to achieve comparable performance across all realizations of the same size. However, \Cref{thm:RG_TF_bound} also presents an issue of discriminability; with high probability, the action of a GNN trained on a large random graph will look very similar to the action of that same GNN applied to a different random graph of the same family. While this may appear to be an artifact of the choice to consider normalized GSOs, we remark that these are used almost exclusively in practice for large graphs. Indeed, without normalization the graph frequency response of a filter can become unbounded as $n$ grows.

\subsection{Transferability of large random graphs}\label{sec:RG_tf}

To make the results in \Cref{prop:RG_similarity} and \Cref{thm:RG_TF_bound} concrete, in this section we will explore specific families of random graphs that satisfy \ref{ass:zero_multiplicity}-\ref{ass:concentration}. To this end, and in order to motivate the results pertaining to concentration of eigenvalues, we will introduce distributions on the spectra of random graphs.

\begin{definition}[Empirical spectral distribution]
The \emph{empirical spectral distribution} (ESD) of a real symmetric matrix $X$, $\mu_X$, is the atomic distribution that assigns equal mass to each of the eigenvalues of $X$, i.e.,
\begin{align}\label{eq:empirical_spectral_distribution}
    \mu_X \coloneqq \frac{1}{n}\sum_{i=1}^n \delta_{\lambda_i(X)}.
\end{align}
\end{definition}

\noindent Since we wish to study the spectra of random graphs, in our setting the ESD will be a \emph{random measure}, i.e., a random variable in the space of probability measures on $\R$ (see \cite{tao2012topics} for more details). We will also study a common limiting distribution of the ESD of random matrices, called Wigner's semicircle law.

\begin{definition}[Wigner's semicircle law]
\emph{Wigner's semicircle law} $\mu$ is the distribution with density defined on $[-1,1]$ as
\begin{align}
    \mu(x) \coloneqq \frac{2}{\pi}\sqrt{1 - x^2},
\end{align}
and zero otherwise. 
\end{definition}

\noindent Wigner's semicircle law is to random matrices what the central limit theorem is to random variables. A result of note, called Wigner's eigenvalue rigidity theorem \cite[Theorem 2.2]{erdHos2012rigidity}, states that the eigenvalues of a matrix whose ESD converges to the semicricle law concentrate around specific distinct values, i.e., \ref{ass:concentration} is satisfied. The exact characterization of the general class of matrices whose empirical spectral distributions converge to Wigner's semicircle law is beyond the scope of this paper (see, e.g, \cite{erdHos2012rigidity, zhu2020graphon}). However, many graphs of interest have corresponding shift operators that satisfy these conditions, including the Erd\"os-R\'enyi and Chung-Lu models, as well as random power law graphs with large enough minimum degree. Indeed, in \cite[Theorem 6]{chung2004spectra} it was shown that the eigenvalues of the normalized Laplacian of any random graph with an appropriate given expected degree distribution converge to Wigner's semicircle law. Moreover, \cite[Theorem 5]{chung2004spectra} provides bounds which may be used to show \ref{ass:spectral_gap} is satisfied when the minimum degree is large enough. Furthermore, in \cite{chakrabarty2021spectra} it was shown that inhomogeneous Erd\"os-R\'enyi random graphs (where the connection probabilities may all be different) admit limiting distributions that are related to Wigner's semicircle law, although they may be difficult to compute in general. The implications of these results in the context of transferability of GNNs are summarized below.


\begin{corollary}\label{corr:RG_transferability}
Let $\{\cG_n\}_{n=1}^\infty$ and $\{\tilde{\cG}_n\}_{n=1}^\infty$ be two independent sequences of almost surely connected random graphs drawn from the same distribution with given expected degrees, with normalized Laplacians $\{S_n\}_{n=1}^\infty$ and $\{\tS_n\}_{n=1}^\infty$, respectively. If the minimum and average degrees are large enough, then for a sequence of GNNs $\{\Phi(\cdot; S_n,\Theta)\}_{n=1}^\infty$ following the conditions of \Cref{thm:RG_TF_bound},
\begin{align*}
    \normed{\Phi(\cdot; S_n,\Theta) - \Phi(\cdot; \tS_n,\Theta)}_\cP \to 0,
\end{align*}
in probability as $n \to \infty$.
\end{corollary}
\begin{proof}
Result of \cite[Theorems 5 and 6]{chung2004spectra}, \cite[Theorem 2.2]{erdHos2012rigidity} and \Cref{thm:RG_TF_bound}.
\end{proof}

It may be possible to achieve almost sure convergence above, depending on the family of random graphs under consideration \cite{tao2012topics, erdHos2012rigidity}. While this result is asymptotic in nature, it is possible to obtain convergence rates which may be informative for finite $n$. Concretely, in the context of \Cref{prop:RG_similarity}, the eigenvalue rigidity result in \cite[Theorem 2.2]{erdHos2012rigidity} suggests that for some fixed $n$, $\epsilon \approx O(n^{-1})$ and $\delta \approx O(n^{-c})$ for some small $c$. In other words, GNNs trained on GSOs of size $n$ satisfying the conditions of \Cref{corr:RG_transferability} will have transferability bounds of the order $1/n$ with high probability.

We remark that these results also apply to the transferability of large random hypergraphs. Consider a random hypergraph with $n$ nodes and $m$ hyperedges where the expected degrees of the hyperedges, i.e., the number of other hyperedges that share at least one node, are given by some fixed $w_1,\ldots,w_m$ and the expected cardinality of the hyperedges are $k_1,\ldots,k_m$. In this case both the clique expansion and line graph will be random with given expected degrees and, hence, we may apply \Cref{corr:RG_transferability} to assert with high probability that the \hgnn~trained using one such random hypergraph will be transferable to others of the same size with a bound of the order $1/n$.

\subsection{Graphon Transferability}\label{sec:graphon_tf}

A \emph{graphon} is a symmetric, measurable function (sometimes called a \emph{kernel}) $W: [0,1]^2 \to [0,1]$ which can be understood as the limit of a sequence of dense undirected graphs where the node index is a continuous set \cite{lovasz2012large}. While many authors explore transferability for sequences of graphs that converge to the same graphon \cite{maskey2021transferability, ruiz2020graphon}, we will focus on the particular related case of random graphs sampled from graphons. This allows us to make use of existing results which explore the spectrum of the normalized Laplacian of such sampled graphs \cite{vizuete2021laplacian}; however, it may be possible to obtain more general results to measure the spectral similarity of sequences of GSOs obtained from graphs converging to the same graphon.

We assume that the graphon is bounded away from zero, so that $\inf_{x,y\in[0,1]}W(x,y) > 0$. In order to generate a random graph with $n$ nodes via the graphon $W$, following \cite{vizuete2021laplacian} we sample some points $x_1,\ldots,x_n \sim \text{Uniform}([0, 1])$ and create the edge $(i,j)$ with a probability $W(x_{(i)}, x_{(j)})$, where $x_{(i)}$ denotes the $i$-th order statistic of the sampled points. We say the resulting graph $\cG_n$ is sampled from the graphon $W$. There is an explicit connection to the results of \Cref{sec:random_transferability}, as it has been shown that graphs sampled from graphons in this manner have fixed expected degree distributions \cite{vizuete2021laplacian}. Indeed, \cite[Lemma 3]{vizuete2021laplacian} shows that the eigenvalues of these sampled graphs concentrate, and \cite[Proposition 3]{vizuete2021laplacian} bounds the spectral gap. Moreover, the graphon being bounded away from zero ensures the sampled graphs will be almost surely connected if they are large enough. Hence, all conditions of \Cref{prop:RG_similarity} are satisfied. With this in mind, we present our result on the transferability of sequences of random graphs sampled from the same graphon.

\begin{corollary}\label{corr:graphon_transferability}
Let $\{\cG_n\}_{n=1}^\infty$ and $\{\tilde{\cG}_n\}_{n=1}^\infty$ be two independent sequences of almost surely connected graphs sampled from the same graphon $W$, which is bounded away from zero, with normalized Laplacians $\{S_n\}_{n=1}^\infty$ and $\{\tS_n\}_{n=1}^\infty$. Then, for a sequence of GNNs $\{\Phi(\cdot; S_n,\Theta)\}_{n=1}^\infty$ following the conditions of \Cref{thm:RG_TF_bound},
\begin{align*}
    \normed{\Phi(\cdot; S_n,\Theta) - \Phi(\cdot; \tS_n,\Theta)}_\cP \to 0,
\end{align*}
in probability as $n \to \infty$.
\end{corollary}
\begin{proof}
Direct result of \cite[Lemma 3 and Proposition 3]{vizuete2021laplacian} and \Cref{thm:RG_TF_bound}.
\end{proof}

\section{Stability of Graph Neural Networks via spectral similarity}\label{sec:stability}

Stability is a generalization property of graph learning architectures that is closely related to transferability. In particular, an architecture with the stability property is one for which small changes in the underlying topology or structure of the graph have a small effect on the way signals are processed. Hence, stability can be understood as a special case of transferability, where one graph is a perturbed version of another graph. However, stability is no less important, as it characterizes the robustness of a graph learning architecture to small changes in the underlying graph, which could arise for example via measurement noise. In the following sections, we will explore stability of graph neural networks by computing the spectral similarity of a graph and its perturbed version. By applying the results from \Cref{sec:transferability} together with the similarity bounds we produce herein, we will show that graph neural networks can be stable to perturbations of the graph structure. In contrast to prior results \cite{gama2020stability}, the results herein along with \Cref{sec:RG_tf} show that stability does not decay as the size of the graph grows.

\subsection{Spectral similarity of perturbed matrices}\label{sec:perturbation_similarity}

To investigate stability, we are interested in small changes to some graph $\cG$ that results in a perturbed version which we call $\tilde{\cG}$. In particular, since we are processing graph signals, we will focus on the graph shift operator $S$ of the original graph $\cG$ and the GSO $\tS$ of the perturbed graph $\tilde{\cG}$. We will explore two types of perturbations, both relative and additive, in order to model how a graph may be changed slightly.

Fundamentally, graph neural networks are spectral operators; the graph frequency response of a filter \eqref{eq:frequency_response} makes this relationship explicit. Well-known results such as Weyl's inequality or the Weilandt-Hoffman inequalities \cite{tao2012topics} show that that the spectrum of a real symmetric matrix is stable to small perturbations. Thus, it stands to reason that if small perturbations of a graph lead to small perturbations in the eigenvalues, then these small perturbations should not change the graph frequency response of a filter very much. To this end, we will explore the spectral similarity of a graph and its perturbed version. In doing so, we can bound the change in the eigenvalues of a graph after a perturbation has been applied.

We study two types of perturbations herein, called \emph{additive} and \emph{relative} perturbations. The first and simplest type involves adding a small perturbation to the GSO $S$ of the original graph $\cG$, so that the GSO $\tS$ of the perturbed graph $\tilde{\cG}$ becomes
\begin{align}
    \tS = S + E.
\end{align}
The perturbation is small in the sense of the operator norm, so that $\normed{E}_{op} \leq \delta$. Such perturbations can be understood as adding or removing edge weight regardless of the original magnitude of the edges, and could be as extreme as adding or removing edges entirely. The other type of perturbation instead affects the edge weights in a manner that is relative to their magnitude, so that
\begin{align}
    \tS = S + \frac{1}{2}(SE + ES),
\end{align}
again with $\normed{E}_{op} \leq \delta$ for some small $\delta > 0$. In both cases we assume that the perturbed GSO $\tS$ will be positive semi-definite. We explore the spectral similarity between the original GSOs and their perturbed versions in the following results, beginning with relative perturbations below.

\begin{proposition}\label{prop:relative_perturbation_similarity}
Given a symmetric, positive semi-definite GSO $S$ and its symmetric, positive semi-definite relatively perturbed version $\tS = S + \frac{1}{2}(SE + ES)$, where $E$ is diagonalizable with $\normed{E}_{op} \leq \delta$, the matrices are 
$\delta$-spectrally similar.
\end{proposition}
\begin{proof}
See \Cref{app:pf_relative_perturbation_similarity}.
\end{proof}

\noindent Consider a relative perturbation via dilation where $E = \delta I$ and thus $\normed{E} = \delta$, and $\tS = (1+\delta)S$. We thus observe that $\delta$-similarity is tight, i.e., for a \emph{general} $E$ with $\normed{E} \leq \delta$, we cannot hope for a better bound on the similarity coefficient. Since our results require spectral similarity to hold regardless of the structure of the perturbation $E$, the bound in \Cref{prop:relative_perturbation_similarity} is thus tight. However, for some arbitrary relative perturbation, it may be the case that a spectral similarity coefficient $\epsilon < \delta$ suffices. Thus, this result and those that follow should be viewed as tight \emph{worst-case} bounds on the coefficient of spectral similarity for general relative perturbations. Next, let us explore additive perturbations.

\begin{proposition}\label{prop:additive_perturbation_similarity}
Given a symmetric, positive semi-definite GSO $S$ and its symmetric, positive semi-definite additively perturbed version $\tS = S + E$, where $\normed{E}_{op} \leq \delta$, if $\text{ker}(E) \subseteq \text{ker}(S)$ then the matrices are $(\delta / \bar{\lambda}(S))$-spectrally similar, where $\bar{\lambda}(S)$ is the smallest non-zero eigenvalue of $S$.
\end{proposition}
\begin{proof}
See \Cref{app:pf_additive_perturbation_similarity}.
\end{proof}

\noindent The condition $\text{ker}(E) \subseteq \text{ker}(S)$ mirrors \ref{ass:zero_multiplicity} from \Cref{prop:RG_similarity}. If $S$ is the normalized Laplacian, this condition enforces that the perturbation not cause the graph to become disconnected, which is reasonable in practice. Indeed, from a graph signal processing perspective, disconnected components have no effect on each other. Further, similarly to \ref{ass:spectral_gap}, if $S$ is the normalized Laplacian then Cheeger's inequality states $h^2_\cG/2 \leq \bar{\lambda}(S) \leq 2h_\cG$, where $h_\cG$ is the Cheeger constant (or conductance) of the graph $\cG$ \cite{chung1997spectral}. Hence, our result in \Cref{prop:additive_perturbation_similarity} tells us that graphs with higher Cheeger constant, i.e., better-connected graphs, will be more similar to their additively perturbed versions. In other words, graphs which are well-connected will result in trained GNNs that exhibit better stability properties than more sparsely connected graphs.

In practice, perturbations to graphs can be modeled as both relative and additive perturbations. Thus, below we consider spectral similarity when both perturbations are present.

\begin{proposition}\label{prop:perturbation_similarity}
Given a symmetric, positive semi-definite GSO $S$, an additive perturbation matrix $D$ such that $\normed{D}_{op} \leq \delta_A$ and $\text{ker}(D) \subseteq \text{ker}(S)$, a diagonalizable relative perturbation matrix $E$ such that $\normed{E}_{op} \leq \delta_R$, and the symmetric, positive semi-definite perturbed GSO $\tS = S + \frac{1}{2}(SE + ES) + D$, the matrices are $(\delta_R + \delta_A/\bar{\lambda}(S))$-spectrally similar.
\end{proposition}
\begin{proof}
Combination of \Cref{prop:relative_perturbation_similarity} and \Cref{prop:additive_perturbation_similarity}.
\end{proof}

While our bounds on spectral similarity are tight for relative perturbations, they are maximal in the sense that we require them to hold for all eigenvalues, i.e., they are equivalent to $(1-\epsilon)\lambda_i(S) \leq \lambda_i(\tS) \leq (1+\epsilon)\lambda_i(S)$ for all $i\in\{1,\ldots,n\}$. However, perturbations in practice will not affect the eigenvalues in such a uniform manner, and thus the following stability bounds may not be tight for arbitrary perturbations. Moreover, for additive perturbations, in \Cref{prop:additive_perturbation_similarity} the quantity $1/\bar{\lambda}(S)$ arises due to the necessity of satisfying spectral similarity for all signals $\bx\in\R^n$, including the eigenvector $\bar{\mathbf{v}}$ associated with the smallest non-zero eigenvalue of $S$. If the signals to be used in practice are not well-aligned with $\bar{\mathbf{v}}$, we may be able to tighten this bound.

\subsection{Stability via spectral similarity}

To achieve our main result on stability of graph neural networks, we combine the above results on perturbations with the transferability bounds from \Cref{sec:transferability}.

\begin{theorem}
Consider a symmetric, positive semi-definite GSO $S$ and its perturbed version $\tS = S + \frac{1}{2}(SE + ES) + D$, where $D$ is an additive perturbation matrix such that $\normed{D}_{op} \leq \delta_A$ and $\text{ker}(D) \subseteq \text{ker}(S)$, and $E$ is a diagonalizable relative perturbation matrix such that $\normed{E}_{op} \leq \delta_R$. The GNN $\Phi(\cdot; S,\Theta)$ with normalized Lipschitz nonlinearities and $L$ layers with $f$ features, each with filters that have unit operator norm and are $C$-integral Lipschitz, satisfies
\begin{align*}
    &\normed{\Phi(\cdot; S,\Theta) - \Phi(\cdot; \tS,\Theta)} \\
    &\quad\leq C L f^L (\delta_R + \delta_A/\bar{\lambda}(S)) + \cO((\delta_R + \delta_A)^2)
\end{align*}
\end{theorem}
\begin{proof}
Follows from \Cref{thm:GNN_bound} and \Cref{prop:perturbation_similarity}.
\end{proof}

In contrast to previous results on the stability of graph neural networks \cite{gama2020stability}, this bound has no dependence on the size of the graph $n$. This can be explained by the effect of the eigenvector differences which appear in the result in \cite{gama2020stability}. Via spectral similarity, we instead upper-bound the difference between the matrices (in the sense of the positive semi-definite cone), i.e., we consider $(1+\epsilon)S$ instead of $\tS$. In this way we can bound the transferability error by considering the eigenvalues, which are known to be stable to perturbations, without considering any changes whatsoever to the eigenvectors, which are known to be unstable \cite{tao2012topics}. Indeed, the results of \Cref{sec:random_transferability} suggest that stability in fact \emph{improves} as the number of nodes $n$ grows larger. Thus, stability is entirely characterized by parameters of the architecture (the integral Lipschitz constant and number of layers), the structure of the original graph (via the smallest nonzero eigenvalue of $S$), and the magnitude of the perturbations that occur. Naturally, larger magnitude perturbations result in looser stability bounds, as does a larger integral Lipschitz constant $C$ and more layers $L$ or features $f$. However, larger $C$, $L$, and $f$ suggest enhanced discriminability, since the GNN can produce sharper filters as $C$ is larger, and more of those filters can be composed as $L$ and $f$ grow. A larger value of $\bar{\lambda}(S)$ when $S$ is the normalized Laplacian suggests better connectedness and, hence, more local smoothing, which explains how it tightens the stability bound. Together, these insights pose a tradeoff between stability and discriminability; if a GNN is indifferent to large perturbations in a graph's structure, it cannot also tell very similar graphs apart.

\section{Simulations}

\begin{table}[ht]
    \centering
    \caption{Cross-validated comparison \& ablation test.}\label{tab:gnn_ablation}
    \begin{tabular}{|l|c|c|}
        \hline
        Model & Validation accuracy & Test accuracy \\
        \hline
        Clique Expansion & $0.631 \pm 0.027$ & $0.552 \pm 0.016$ \\
        \hline
        Line Graph & $0.625 \pm 0.057$ & $0.588 \pm 0.016$ \\
        \hline
        HGNN \cite{feng2019hypergraph} & $0.638 \pm 0.024$ & $0.580 \pm 0.012$ \\
        \hline
        \hgnn~(ours) & $\mathbf{0.889 \pm 0.155}$ & $\mathbf{0.852 \pm 0.123}$ \\
        \hline
    \end{tabular}
\end{table}

To evaluate the performance of \hgnn, we train several graph and hypergraph neural networks to solve a hyperedge source localization problem, wherein the goal is to determine the source hyperedge of a diffusion process that has been occurring over the nodes of a hypergraph for an unknown period of time. Hence, the input is a node signal, but the desired output is a hyperedge signal. This nonlinear diffusion of both node and edge signals occurs via the hypergraph Laplacian \eqref{eq:hg_laplacian}. To make the problem more difficult, we add both input and measurement noise to the resulting signals, so our problem also involves hypergraph signal de-noising. We perform an ablation test for \hgnn~by comparing with GNNs that use only the clique expansion or line graph of the original hypergraph, as well as a comparison with HGNN \cite{feng2019hypergraph}, showing the improvement in performance of \hgnn~relative to approaches which diffuse the signal across \emph{only} the nodes or hyperedges. Since none of these other approaches can natively handle both node and hyperedge signals, where appropriate we pool all signals according to hyperedge inclusions, in the same manner as \hgnn. The full experimental setup is detailed in \Cref{app:experiments}. In particular, we note that all methods (including HGNN) were constrained to have normalized and integral Lipschitz filter weights to match the setting of \Cref{thm:GNN_bound}.

The results of this study are presented in \Cref{tab:gnn_ablation}. Mean and standard deviations of performance metrics were computed after randomly shuffling the data 5 times. Hyperparameters were set based on the highest upper-confidence bound cross-validation score (mean plus standard deviation). \hgnn~exhibits a $45-54\%$ improvement in test accuracy over GNNs which use only the clique expansion or line graph, illustrating the value of combining these representations for problems that inherently involve a nonlinear diffusion process across both the nodes and hyperedges of a hypergraph. Moreover, it shows a $47\%$ improvement over HGNN when both methods have normalized and integral Lipschitz filter weights, suggesting superior performance when transferability is desired.

\appendix
\interdisplaylinepenalty=100

\subsection{Transferability bounds via spectral similarity}\label{app:pf_spectral_similarity_bound}
\begin{proof}[Proof of \Cref{prop:spectral_similarity_bound}]
By spectral similarity, we have that
\begin{align}
    &(1-\epsilon)S \preceq \tS \preceq (1+\epsilon)S \\
    &\Rightarrow~ (1-\epsilon)\bx^\intercal S \bx \leq \bx^\intercal \tS \bx \leq (1+\epsilon)\bx^\intercal S \bx \quad\forall \bx\in\R^n.
\end{align}
Hence,
\begin{align*}
    H(\tS) - H(S) &= \sum_{k=0}^\infty h_k(\tS^k - S^k) = \sum_{k=1}^\infty h_k(\tS^k - S^k) \\
    &\preceq \sum_{k=1}^\infty h_k(((1+\epsilon)S)^k - S^k).
\end{align*}
For symmetric matrices $\normed{A}_{op} \leq \normed{B}_{op}$ if $A \preceq B$. Thus, using the first-order expansion $((1+\epsilon)S)^k = (1+k\epsilon)S^k + O(\epsilon^2)$, since graph filters are permutation equivariant we have
\begin{align*}
    \normed{H(\tS) - H(S)}_{\cP} &= \normed{H(\tS) - H(S)}_{op} \\
    &\leq \normed{\sum_{k=1}^\infty h_k(((1+\epsilon)S)^k - S^k)}_{op} \\
    &= \normed{\sum_{k=1}^\infty h_k((1+k\epsilon)S^k - S^k) + O(\epsilon^2)}_{op} \\
    &\leq \normed{\sum_{k=1}^\infty h_k k\epsilon S^k}_{op} + \cO(\epsilon^2).
\end{align*}
Next, notice that the derivative of the frequency response is $\th'(\lambda) = \sum_{k=1}^\infty h_k k\lambda^{k-1}$. Thus, using the graph Fourier representation of $\bx$,
\begin{align*}
    \sum_{k=1}^\infty h_k k\epsilon S^k \bx &= \epsilon\sum_{k=1}^\infty h_k k S^k \left(\sum_{i=1}^n\tx_i \bv_i \right) \\
    &= \epsilon\sum_{i=1}^n\tx_i\sum_{k=1}^\infty h_k kS^k \bv_i \\
    &= \epsilon\sum_{i=1}^n\tx_i\sum_{k=1}^\infty h_k k \lambda_i^k \bv_i \\
    &= \epsilon\sum_{i=1}^n\tx_i \lambda_i\th'(\lambda_i) \bv_i.
\end{align*}
By the integral Lipschitz assumption of the filter, $\lambda\th'(\lambda) \leq C$. Moreover, since the signal is assumed to have a unit norm, $\normed{\bx}_2 = \normed{\tilde{\bx}}_2 = 1$. Thus, by orthonormality of the $\bv_i$,
\begin{align*}
    \normed{\sum_{k=1}^\infty h_k k\epsilon S^k \bx}_2^2 &= \normed{\epsilon\sum_{i=1}^n\tx_i \lambda_i\th'(\lambda_i) \bv_i}_2^2 \\
    &= \epsilon^2\sum_{i=1}^n(\tx_i\lambda_i\th'(\lambda_i))^2 \\
    &\leq (C\epsilon)^2\sum_{i=1}^n\tx_i^2 = (C\epsilon)^2.
\end{align*}
Finally, we thus have
\begin{align*}
    \normed{H(\tS) \!-\! H(S)}_{\cP} &\!\leq\! \normed{\sum_{k=1}^\infty h_k k\epsilon S^k}_{op} \!+\! \cO(\epsilon^2) \leq C\epsilon \!+\! \cO(\epsilon^2),
\end{align*}
as desired. Moreover, note that if we apply only one shift operation with a bias term, i.e., $H(S) = h_0 I + h_1 S$, then the first-order expansion performed earlier is exact and $\normed{H(\tS) - H(S)}_{\cP} \leq C\epsilon$.
\end{proof}

\subsection{Transferability of GNNs}\label{app:pf_GNN_bound}

\begin{proof}[Proof of \Cref{thm:GNN_bound}]
We wish to bound the quantity $||\Phi(\cdot; S,\Theta) - \Phi(\cdot; \tS,\Theta)||_\cP$, and proceed similarly to \cite[Theorem 4]{gama2020stability}. At an arbitrary layer $l\in\{1,\ldots,L\}$ of $\Phi(\cdot; S,\Theta)$ with $f_l$ features (where $l=L$ is the output layer), the (possibly hidden) node signals are some $\{\bx_l^f\}_{f=1}^{f_l}$, where $\bx_l^f \in \R^{n}$. Similarly, the node signals in the GNN $\Phi(\cdot; \tS,\Theta)$ at layer $l$ are denoted by $\{\tbx_l^f\}_{f=1}^{f_l}$. We thus wish to bound quantities of the form $||\bx_l^f - \tbx_l^f||_2$, i.e.,
\begin{align}\label{eq:pf_GNN_output_bound}
    \normed{\sigma\!\left(\sum_{g=1}^{f_{l-1}} H^{fg}_l(S)\bx_{l-1}^g\right) \!-\! \sigma\!\left(\sum_{g=1}^{f_{l-1}} H^{fg}_l(\tS)\tbx_{l-1}^g\right)}_2\!.
\end{align}
Since the nonlinearities are assumed to be normalized Lipschitz, $|\sigma(x) - \sigma(y)| \leq |x-y|$. So, by the triangle inequality,
\begin{align*}
    &\normed{\bx_l^f - \tbx_l^f}_2 \\
    &\leq \sum_{g=1}^{f_{l-1}} \normed{H^{fg}_l(S)\bx_{l-1}^g - H^{fg}_l(\tS)\tbx_{l-1}^g}_2 \\
    &= \sum_{g=1}^{f_{l-1}} \normed{H^{fg}_l(S)\!\left(\bx_{l-1}^g \!-\! \tbx_{l-1}^g\right) \!+\! \left(H^{fg}_l(S) \!-\! H^{fg}_l(\tS)\right)\!\tbx_{l-1}^g}_2 \\
    &\leq \sum_{g=1}^{f_{l\!-\!1}} \!\normed{H^{fg}_l\!(S)}_{op}\normed{\bx_{l\!-\!1}^g \!-\! \tbx_{l\!-\!1}^g}_2 \!+\! \normed{H^{fg}_l\!(S) \!-\! H^{fg}_l\!(\tS)}_{op}\!\normed{\tbx_{l\!-\!1}^g}_2\!.
\end{align*}
Notice similarly that
\begin{align*}
    \normed{\bx_l^{g_l}}_2 &= \normed{\sigma\left(\sum_{g_{l-1}=1}^{f_{l-1}} H_l^{g_l g_{l-1}}(S)\bx_{l-1}^{g_{l-1}}\right)}_2 \\
        &\leq \sum_{g_{l-1}=1}^{f_{l-1}} \normed{H^{g_l g_{l-1}}_l(S)\bx_{l-1}^{g_{l-1}}}_2 \\
        &\leq \sum_{g_{l-1}=1}^{f_{l-1}} \normed{H^{g_l g_{l-1}}_l(S)}_{op}\normed{\bx_{l-1}^{g_{l-1}}}_2 \\
        &\leq \sum_{g_{l-1}=1}^{f_{l-1}} \cdots \sum_{g_{0}=1}^{f_{0}} \normed{\bx_{0}^{g_{0}}}_2 \prod_{s=1}^{l}\normed{H^{g_s g_{s-1}}_s(S)}_{op}.
\end{align*}
Since the GSOs $S$ and $\tS$ are $\epsilon$-spectrally similar, by \Cref{prop:spectral_similarity_bound} we have $||H^{g_l g_{l-1}}_1(S) - H^{g_l g_{l-1}}_1(\tS)||_{op} \leq C_l^{g_l g_{l-1}}\epsilon + \cO(\epsilon^2)$ for all $g_l$, $g_{l-1}$, where $C_l^{g_l g_{l-1}}$ is the integral Lipschitz constant of the filter in the $l$-th layer for the $g_{l-1}$-th input feature and $g_l$-th output feature. Combining these results,
\begin{align*}
    &\normed{\bx_l^{g_l} - \tbx_l^{g_l}}_2 \\
    &\leq \sum_{g_{l-1}=1}^{f_{l-1}} \normed{H^{g_l g_{l-1}}_l(S)}_{op}\normed{\bx_{l-1}^{g_{l-1}} - \tbx_{l-1}^{g_{l-1}}}_2 \\
    &\qquad\quad+ (C_l^{g_l g_{l-1}}\epsilon + \cO(\epsilon^2)) \normed{\tbx_{l-1}^g}_2 \\
    &\leq \cO(\epsilon^2) + \sum_{g_{l-1}=1}^{f_{l-1}} \normed{H^{g_l g_{l-1}}_l(S)}_{op}\normed{\bx_{l-1}^{g_{l-1}} - \tbx_{l-1}^{g_{l-1}}}_2 \\
    &\qquad\quad+ \!\sum_{g_{l-2}=1}^{f_{l-2}} \!\cdots \sum_{g_{0}=1}^{f_{0}} \normed{\tbx_{0}^{g_{0}}}_2 C_l^{g_l g_{l-1}} \epsilon \prod_{s=1}^{l-1}\normed{H^{g_s g_{s-1}}_s(S)}_{op}\!.
\end{align*}
This establishes a recurrence relation for $\normed{\bx_l^{g_l} - \tbx_l^{g_l}}_2$. Notice that for $l=1$ and any $g_1 \in \{1,\ldots,f_1\}$,
\begin{align*}
    \normed{\bx_1^{g_1} - \tbx_1^{g_1}}_2 \leq \sum_{g_0=1}^{f_{0}} C_1^{g_1 g_0} \epsilon \normed{\bx_{0}^{g_0}}_2 + \cO(\epsilon^2),
\end{align*}
since the input signals to both GNNs are the same and, hence, $\normed{\bx_0^{g_0}}_2 = \normed{\tbx_0^{g_0}}_2$ and $\normed{\bx_{0}^{g_0} - \tbx_{0}^{g_0}}_2 = 0$ for all $g_0 \in \{1,\ldots,f_0\}$. Thus we may solve the recurrence relation above to obtain
\begin{align*}
    &\normed{\bx_l^{g_l} - \tbx_l^{g_l}}_2 \\
    &\!\leq \!\sum_{g_{l\!-\!1}=1}^{f_{l-1}} \!\cdots\! \sum_{g_{0}=1}^{f_{0}} \!\normed{\bx_{0}^{g_{0}}}_2\! \sum_{s=1}^l C_s^{g_s g_{s\!-\!1}} \epsilon \!\prod_{\substack{t=1 \\ t \neq s}}^{l}\normed{H^{g_t g_{t\!-\!1}}_t\!(S)}_{op} \!+\! \cO(\epsilon^2).
\end{align*}
This inequality makes explicit the effects of the integral Lipschitz constants $C_l^{g_l g_{l-1}}$ as well as the operator norms $\normed{H^{g_t g_{t-1}}_t(S)}_{op}$ of each filter, which are functions of both the (learned) filter coefficients and the spectrum of the GSO $S$ used for training. Both quantities control the amplification of the input signals; if they are smaller, the magnitude of the difference in the signals at layer $l$ is smaller. The output signals of the GNNs are the quantities $\{\bx_L^f\}_{f=1}^{f_L}$ and $\{\tbx_L^f\}_{f=1}^{f_L}$. Hence, since we wish to bound the difference in the output of the GNNs, for an arbitrary input signal $\bx_0 = \{\bx_0^f\}_{f=1}^{f_0}$, we have
\begin{align*}
    &\normed{\Phi(\bx_0 ; S,\Theta) - \Phi(\bx_0 ; \tS,\Theta)}_2 \\
    &= \sum_{g_L=0}^{f_L} \normed{\bx_L^{g_L} - \tbx_L^{g_L}}_2 \\
    &\leq \!\sum_{g_L=0}^{f_L} \!\cdots\! \sum_{g_{0}=1}^{f_{0}} \normed{\bx_{0}^{g_{0}}}_2 \sum_{s=1}^L C_{s}^{g_s g_{s-1}}\epsilon \prod_{t=1, t \neq s}^{L}\normed{H^{g_t g_{t-1}}_t(S)}_{op} \\
    &\quad+ \cO(\epsilon^2).
\end{align*}
If the filter responses are bounded, i.e., $\normed{H^{g_t g_{t-1}}_t(S)}_{op} \leq c$ (which can be achieved via normalization during training), all filters share an integral Lipschitz constant $C$ (which is possible by adding a penalty to the training loss), and the input signals have unit norm (via normalization before training), then
\begin{align*}
    \normed{\Phi(\bx_0 ; S,\Theta) - \Phi(\bx_0 ; \tS,\Theta)}_2 &\leq C L c^{L-1} \prod_{s=0}^{L} f_s \epsilon + \cO(\epsilon^2).
\end{align*}
The constant $c$ can be understood as controlling the amplification or contraction of the input and hidden signals, while the number of features $f_s$ and layers $L$ determine how many filters are stacked together, each obeying the bound from \Cref{prop:spectral_similarity_bound}, which depends on the integral Lipschitz constant $C$ and the spectral similarity coefficient $\epsilon$. The desired result is achieved by setting $c=1$ and $f_s=f$ for all $s\in\{0,\ldots,L\}$.
\end{proof}

\subsection{Spectral similarity under relative perturbations}\label{app:pf_relative_perturbation_similarity}
\begin{proof}[Proof of \Cref{prop:relative_perturbation_similarity}]
Recall by definition of $\epsilon$-spectral similarity \eqref{eq:spectral_similarity} that
\begin{align*}
    &(1-\epsilon)S \preceq \tS \preceq (1+\epsilon)S \nonumber \\
    &\Leftrightarrow (1-\epsilon)\bx^\intercal S \bx \leq \bx^\intercal\tS \bx \leq (1+\epsilon)\bx^\intercal S\bx \quad\forall \bx\in\R^n.
\end{align*}
Given some perturbation matrix $E$, we wish to determine the coefficient of spectral similarity, $\epsilon$. Let us begin with the latter inequality,
\begin{align*}
    \tS \preceq (1+\epsilon)S &\Leftrightarrow S + \frac{1}{2}(SE + ES) \preceq (1 + \epsilon)S \\
    &\Leftrightarrow \epsilon S - \frac{1}{2}(SE + ES) \succeq 0 \\
    &\Leftrightarrow \bx^\intercal\!\left(\epsilon S - \frac{1}{2}(SE + ES)\right)\!\bx \geq 0 \quad \forall \bx\in\R^n.
\end{align*}
Note that, since $S$ is PSD, we may always choose $\epsilon$ to be large enough that $\epsilon S - \frac{1}{2}(SE + ES)$ is PSD. Let us diagonalize $E = UMU^\intercal$, and recall that $\normed{E}_{op} = |\lambda_{max}(E)| \leq \delta$. Hence,
\begin{align*}
    &\bx^\intercal\left(\epsilon S - \frac{1}{2}(SE + ES)\right)\bx \\
    &= \bx^\intercal S((\epsilon/2)I - E/2)\bx + \bx^\intercal ((\epsilon/2)I - E/2)S\bx \\
    &= \bx^\intercal SU((\epsilon/2)I \!-\! M/2)U^\intercal \bx + \bx^\intercal U((\epsilon/2)I \!-\! M/2)U^\intercal S\bx \\
    &\geq \frac{1}{2}\bx^\intercal SU(\epsilon I - \delta I)U^\intercal \bx + \frac{1}{2}\bx^\intercal U(\epsilon I - \delta I)U^\intercal S\bx \\
    &= \frac{1}{2}(\epsilon - \delta)\bx^\intercal SUU^\intercal \bx + \frac{1}{2}(\epsilon - \delta)\bx^\intercal UU^\intercal S\bx \\
    &= (\epsilon - \delta)\bx^\intercal S\bx,
\end{align*}
with equality when $E = \delta I$. Since $S$ is PSD, $\bx^\intercal S\bx \geq 0~\forall \bx\in\R^n$. Thus, if $\epsilon \geq \delta$, then $\tS \preceq (1+\epsilon)S$. Moreover, when $E = \delta I$, $\tS = (1 + \delta)S = (1 + \epsilon)S$, and $\delta$-similarity is tight.

Next, notice similarly that
\begin{align*}
    (1-\epsilon)S \preceq \tS &\Leftrightarrow \epsilon S + \frac{1}{2}(SE + ES) \succeq 0 \\
    &\Leftrightarrow \bx^\intercal\left(\epsilon S + \frac{1}{2}(SE + ES)\right)\bx \geq 0 ~\forall \bx\in\R^n.
\end{align*}
Then, as above,
\begin{align*}
    &\bx^\intercal\left(\epsilon S + \frac{1}{2}(SE + ES)\right)\bx \\
    &= \bx^\intercal SU((\epsilon/2)I \!+\! M/2)U^\intercal\bx + \bx^\intercal U((\epsilon/2)I \!+\! M/2)U^\intercal S\bx \\
    &\geq \frac{1}{2}\bx^\intercal SU(\epsilon I + \lambda_{min}(E)I)U^\intercal\bx \\
    &\quad + \frac{1}{2}\bx^\intercal U(\epsilon I + \lambda_{min}(E)I)U^\intercal S\bx \\
    &= (\epsilon + \lambda_{min}(E))\bx^\intercal S\bx.
\end{align*}
Since $|\lambda_{min}(E)| \leq |\lambda_{max}(E)| = \delta$, $\epsilon \geq \delta$ implies $(1-\epsilon)S \preceq \tS$. Hence if $\epsilon \geq \delta$, we have that $(1-\epsilon)S \preceq S + SE + ES \preceq (1+\epsilon)S$.
\end{proof}

\subsection{Spectral similarity under additive perturbations}\label{app:pf_additive_perturbation_similarity}
\begin{proof}[Proof of \Cref{prop:additive_perturbation_similarity}]
As in \Cref{prop:relative_perturbation_similarity}, let us find the value of $\epsilon$ which satisfies the inequality
\begin{align*}
    \tS \preceq (1+\epsilon)S &\Leftrightarrow S + E \preceq (1+\epsilon)S \\
    &\Leftrightarrow \epsilon S - E \succeq 0 \\
    &\Leftrightarrow \bx^\intercal(\epsilon S - E)\bx \geq 0 \quad\forall \bx\in\R^n.
\end{align*}
For any $\bx \in \text{ker}(S)$, $\bx^\intercal(\epsilon S - E)\bx = -\bx^\intercal E\bx$, but by assumption $\text{ker}(E) \subseteq \text{ker}(S)$. Hence, regardless of $\epsilon$, $\bx^\intercal(\epsilon S - E)\bx = 0$ for $\bx \in \text{ker}(S)$. Now, for $\bx\not\in\text{ker}(S)$, since $\normed{E}_{op} \leq \delta$,
\begin{align*}
    \bx^\intercal(\epsilon S - E)\bx &= \epsilon \bx^\intercal S\bx - \bx^\intercal E\bx \\
    &\geq \epsilon \bx^\intercal S\bx - \delta\normed{\bx}_2 \\
    &\geq \epsilon \bar{\lambda}(S)\normed{\bx} - \delta\normed{\bx}_2 \\
    &= (\epsilon\bar{\lambda}(S) - \delta)\normed{\bx}_2,
\end{align*}
where we have used the fact that $\min\{\bx^\intercal S \bx \ | \ \bx \not\in\ker(S)\} = \bar{\lambda}(S)\normed{\bx}_2$. Thus, we must have that $\epsilon \geq \delta / \bar{\lambda}(S)$.  Now, for the second inequality,
\begin{align*}
    (1-\epsilon)S \preceq \tS &\Leftrightarrow \bx^\intercal(\epsilon S + E)\bx \geq 0 \quad\forall \bx\in\R^n.
\end{align*}
Clearly for $\bx\in\text{ker}(S)$, $\bx^\intercal(\epsilon S + E)\bx = 0$. Then, for $\bx\not\in\text{ker}(S)$,
\begin{align*}
    \bx^\intercal(\epsilon S + E)\bx &= \epsilon \bx^\intercal S\bx + \bx^\intercal E\bx \\
    &\geq (\epsilon\bar{\lambda}(S) + \lambda_{min}(E))\normed{\bx}_2.
\end{align*}
Thus, if $\epsilon \geq \max\{-\lambda_{min}(E)/\bar{\lambda}(S), 0\}$ then $(1-\epsilon)S \preceq \tS$. However, $\delta = |\lambda_{max}(E)| \geq |\lambda_{min}(E)|$. Hence, it is sufficient for $\epsilon \geq \delta/\bar{\lambda}(S)$ to ensure $(1-\epsilon)S \preceq S + E \preceq (1+\epsilon)S$.
\end{proof}

\subsection{Spectral similarity of random graphs}\label{app:pf_RG_similarity}

\begin{proof}[Proof of \Cref{prop:RG_similarity}]
For ease of notation, fix some $n$ and set $\lambda_i \coloneqq \lambda_i(S_n)$ and $\tlambda_i \coloneqq \lambda_i(\tS_n)$. Hence, denoting $\cI \coloneqq \left\{i \in \{1,\ldots,n\} \ : \ \lambda_i(S_n) \neq 0\right\}$, note that \ref{ass:zero_multiplicity} implies almost surely that we can order the eigenvalues of $\tS_n$ so that $\cI = \left\{i \in \{1,\ldots,n\} \ : \ \lambda_i(\tS_n) \neq 0\right\}$ and, thus, $|\lambda_i - \tlambda_i| = 0$ for all $i \not\in\cI$. Thus, it suffices to show that $\forall \epsilon > 0, \delta > 0$ there exists some $N$ such that for any $n\geq N$,
\begin{align}
    &P\left((1 - \epsilon)\lambda_i \leq \tlambda_i \leq (1 + \epsilon)\lambda_i~\forall i\in\cI\right) > 1-\delta \nonumber \\
    \Leftrightarrow~& P\left(|\lambda_i - \tlambda_i| \leq \epsilon|\lambda_i|~\forall i\in\cI\right) > 1-\delta.
\end{align}
By \ref{ass:spectral_gap}, the non-trivial eigenvalues of $S_n$ and $\tS_n$ are bounded away from zero by some $c>0$ almost surely. Thus, since $P(A) > P(B)$ if $A \supseteq B$,
\begin{align*}
    &P\left(|\lambda_i - \tlambda_i| \leq \epsilon|\lambda_i|~\forall i\in\cI\right) \\
    &\geq P\left(|\lambda_i - \tlambda_i| \leq  c\epsilon~\forall i\in\cI\right) \\
    &= P\left(|\lambda_i - \gamma_i - (\tlambda_i - \gamma_i)| \leq c\epsilon~\forall i\in\cI\right) \\
    &\geq P\left(|\lambda_i - \gamma_i| + |\tlambda_i - \gamma_i| \leq c\epsilon~\forall i\in\cI\right) \\
    &\geq P\Big(|\lambda_i - \gamma_i| \leq c\epsilon/2~\forall i\in\cI\Big)P\!\left(|\tlambda_i - \gamma_i| \leq c\epsilon/2~\forall i\in\cI\right)\!,
\end{align*}
where the last line follows by independence. Then by \ref{ass:concentration}, for any $\epsilon > 0$, $\delta > 0$ we can choose $N_1$ large enough such that $P(|\lambda_i - \gamma_i| \leq c\epsilon/2~\forall i\in\cI) > \sqrt{1-\delta}$ (and similarly choose $N_2$ for all $\tlambda_i$), and the result follows by picking $N = \max\{N_1, N_2\}$.
\end{proof}

\subsection{Experimental setup}\label{app:experiments}

The hypergraph is created by randomly sampling 500 points on a 3-dimensional torus with inner radius $1$ and outer radius $2$, and keeping maximal hyperedges of a Vietoris-Rips complex \cite{zomorodian2010fast} with radius $0.4$, i.e., constructing simplices between any points jointly within a 2-norm distance $0.4$. $10$ hyperedges are selected at random to be the sources, making the source localization problem a $10$-class classification problem. 

To generate $K$ data samples $\{(\bx_k,y_k)\}_{k=1}^K$, we pick $10$ hyperedges at random to be the sources. For each source hyperedge $i$, we construct a node source signal for node $j$:
\begin{align*}
    x_{0,j}^i = \begin{cases}
            1 + z, & \text{node $i$ is in the source hyperedge}, \\
            z, & \text{otherwise.}
        \end{cases}
\end{align*}
where $z \sim \text{Normal}(0, 0.01)$ is independent Gaussian white noise. This noisy source signal is diffused for $t_{max}=30$ time steps via the nonlinear hypergraph Laplacian \eqref{eq:hg_laplacian} to generate the sequence $\{\bx_t^i\}_{t=0}^{t_{max}}$. Samples are then of the form $(\bx_t^i + \bz, y_i)$, where we pick $\bx_t^i$ by randomly sampling a source hyperedge $i$ and time $t$ and add measurement noise $\bz \sim \text{Normal}(\mathbf{0}, 0.01I_n)$, and the hyperedge label is $y_i = i$. We use $500$ of these signals for training (including cross-validation), and $300$ of these signals for testing. Note that while we may sample a signal from the same source at the same time more than once, they will not be identical due to the added independent measurement noise.

All GNNs have two graph filtering layers with one input feature per node and a fixed readout layer to select signals from the $10$ candidate source hyperedges. The filters are normalized during training to ensure $|h(\lambda)| \leq 1$ and the integral Lipschitz constant is constrained to be less than $10$ via a loss penalty term. Pooling from node to hyperedge signals is done based on node inclusion in the hyperedges. We train using $\texttt{adam}$ with weights $0.9$ and $0.999$ with a learning rate of $0.0005$, decay rate of $0.99$ and decay period of $20$.
These hyperparameters, as well as others such as the number of filters and number of filter taps were chosen via 5-fold cross-validation.

\bibliographystyle{plain}
\bibliography{references, references_2}

\end{document}